\documentclass[11pt,a4paper]{article}
\usepackage[hyperref]{acl2019}
\usepackage{times}
\usepackage{latexsym}
\usepackage{amsmath, amssymb, graphicx, bm, multirow}
\usepackage{url}
\usepackage{todonotes}
\usepackage{subfigure}

\aclfinalcopy 

\title{Time Masking: Leveraging Temporal Information\\ in Spoken Dialogue Systems}

\author{Rylan Conway and Lambert Mathias\\
Amazon Alexa AI \\
{\tt \{conrylan, mathiasl\}@amazon.com} \\}

\date{}

\begin{document}
\maketitle

\setlength\abovedisplayskip{6pt}
\setlength\belowdisplayskip{6pt}

\begin{abstract}
In a spoken dialogue system, dialogue state tracker (DST) components track the state of the conversation by updating a distribution of values associated with each of the slots being tracked for the current user turn, using the interactions until then.
Much of the previous work has relied on modeling the natural order of the conversation, using distance based offsets as an approximation of time.
In this work, we hypothesize that leveraging the wall-clock temporal difference between turns is crucial for finer-grained control of dialogue scenarios.
We develop a novel approach that applies a {\it time mask}, based on the wall-clock time difference, to the associated slot embeddings and empirically demonstrate that our proposed approach outperforms existing approaches that leverage distance offsets, on both an internal benchmark dataset as well as DSTC2.

\end{abstract}

\section{Introduction}
\label{sec:intro}
Modern spoken dialogue systems -- such as Intelligent Personal Digital Assistants (IPDAs) like Google Assistant, Siri, and Alexa -- provide users a natural language interface to help complete tasks such as reserving restaurants, checking the weather, playing music etc.
Spoken language understanding (SLU) is a central component in such dialogue systems, and is responsible for parsing the natural language text to semantic frames.
In task-oriented spoken dialogue systems, a key challenge is tracking entities the user introduced in previous dialogue turns.
For example, if a user request for \textit{what's the weather in arlington} is followed by \textit{how about tomorrow}, the dialogue system has to keep track of the entity \textit{arlington} being referenced.
Typically, this is formulated as a dialogue state tracking (DST) task~\citep{henderson2014word, mrkvsic2016neural}.

Previous approaches to dialogue state tracking have mostly focused on dialogue representations~\citep{mrkvsic2016neural}, dealing with noisy input~\citep{henderson2012discriminative, mesnil2015using}, or tracking slots from multiple domains~\citep{henderson2014word, rastogi2017scalable, cc-inter}.
In this paper, we focus on temporal information associated with each dialogue turn.
Although the dialogue representations -- typically encoded using LSTMs -- are able to implicitly capture the temporal order in the sequence of dialogue turns, we hypothesize that explicitly and accurately encoding temporal information is essential for resolving ambiguity in dialogue state tracking.
Recently,~\citep{cc-inter} presented work that models  the slot distance offset from the current turn using a one-hot representation input to the DST module.
Alternatively, ~\citep{su2018time} leverage the distance offset in an attention mechanism.
We posit that the notion of time based on distance offset relative to the current turn is too coarse-grained and often insufficient for resolving ambiguities associated with more complex multi-turn dialogues.
For example, in a dialogue \textit{``how far is issaquah?''} followed by \textit{``what is the weather like?"} we could have two possible interpretations -- a follow-up utterance issued within 10  seconds would indicate that the user is referring to the city slot of ``Issaquah" from the previous turn, whereas, if the follow-up utterance is more than 30 seconds apart there is a good chance that the user was just inquiring about the weather in their current location.
In this case, a dialogue system that only encodes the distance offset will be unable to correctly disambiguate the aforementioned situation.
Based on this intuition, we develop a novel approach for incorporating temporal information in dialogue state tracking by using a time mask over the slots. 

To summarize, we introduce the notion of a \textit{time mask} to incorporate temporal information into the embedding for slots.
In contrast to previous approaches using distance offsets, we propose leveraging the wall-clock time difference between the current turn and the previous turns in the dialogue to explicitly model temporal information.
Furthermore, we demonstrate how domain and intent information can be mixed in with the temporal information in this framework to improve DST accuracy.
We demonstrate empirically that our proposed approach improves over the baseline that only encodes distance offsets as temporal information.

\section{Approach}
\label{sec:approach}
\subsection{Slot Carryover Task Description}
\label{ssec:sc}
In this paper, we build on the approach in~\citep{cc-inter}. For completeness, we define the carryover task formulation here, but refer readers to the original work for architecture details.  A dialogue turn at time $t$ is defined as the tuple $\{a_t, \bm{S}_t, \bm{w}_t\}$, where $\bm{w_t} \in \mathcal{W}$ is  a sequence of words $\{w_{it}\}_{i=1}^{N_t}$;
$a_t \in \mathcal{A}$ is the dialogue act; and $\bm{S}_t$ is a set of slots, where each slot $s$ is a key-value pair $s=\{k,v\}$, with $k\in \mathcal{K}$ being the slot name (or slot key), 
and $v\in \mathcal{V}$ being the slot value. A user turn is represented by $\bm{u}_t=\{a_t^u, \bm{S}_t^u, \bm{w}_t^u\}$ and a system turn is represented by $\bm{v}_t=\{a_t^v, \bm{S}_t^v, \bm{w}_t^v\}$. Given a sequence of $D$ user turns $\{\bm{u}_{t-D}, \dots, \bm{u}_{t-2}, \bm{u}_{t-1}\}$;
and their associated system turns $\{\bm{v}_{t-D}, \dots, \bm{v}_{t-2}, \bm{v}_{t-1}\}$\footnote{For simplicity we assume a turn taking model - a user turn and system turn alternate.};
and the current user turn $\bm{u}_t$, we  construct a candidate set of slots from the context as 
\begin{eqnarray}
    C(\bm{S})=\bigcup\limits_{\substack{j=t-D \\ i \in {u,v}}}^{t} \bm{S}^i_j.
\end{eqnarray}
For a candidate slot $s \in C(\bm{S})$, for the dialogue turn at time $t$, the probability to carryover the slot is defined as
\begin{eqnarray}
P(+|s, d(s), \bm{u}_t, \bm{u}_{t-D}^{t-1}, \bm{v}_{t-D}^{t-1}),
\end{eqnarray}
where $d(s) \in [0, D]$ is an integer value describing the offset of the candidate slot from the current turn $\bm{u}_t$.
The final carryover decision is determined by comparing the carryover probability to a tunable decision threshold

\begin{figure*}[ht]
\begin{center}
\includegraphics[scale=0.12]{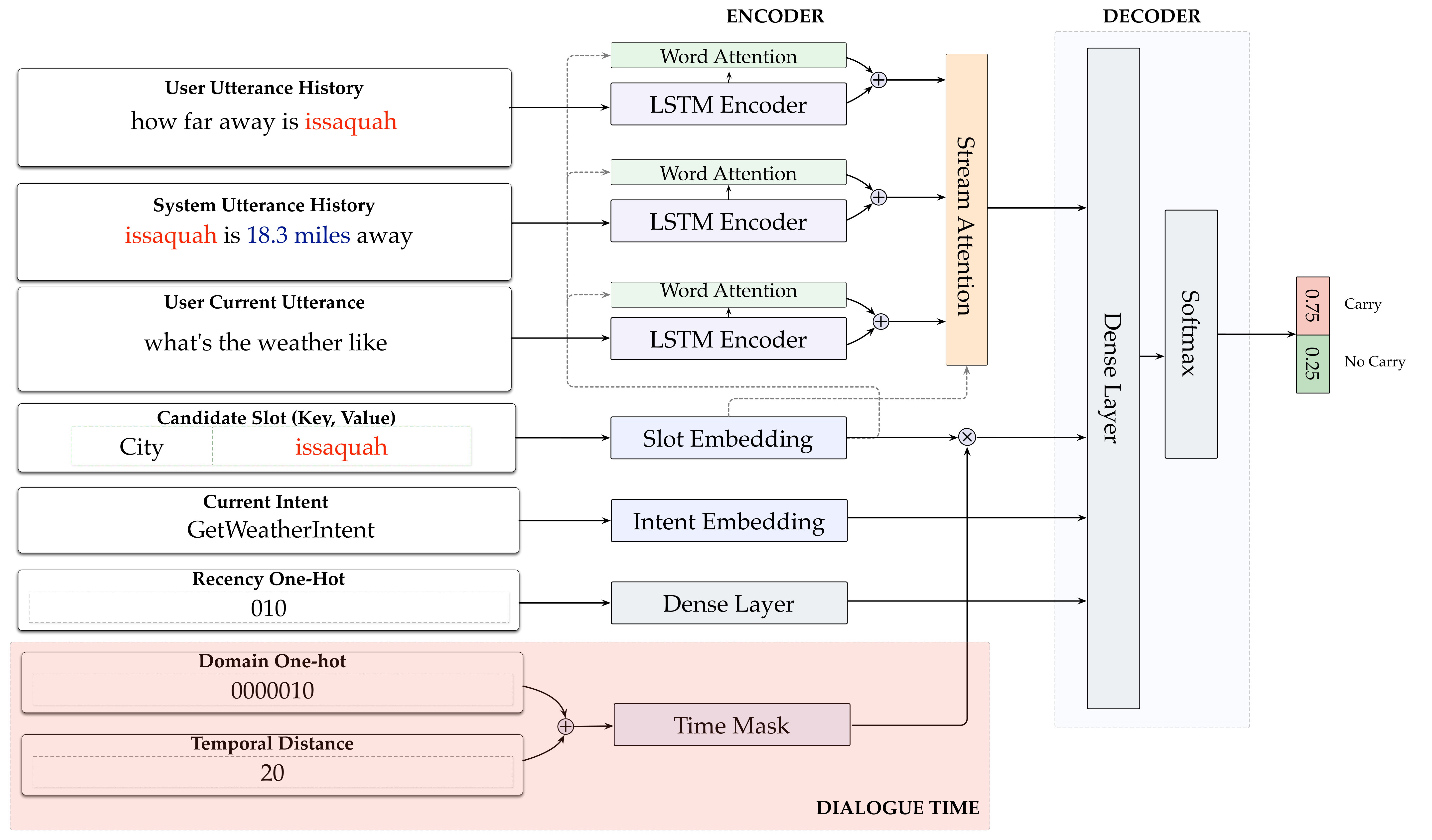}
\caption{Slot carryover architecture from~\cite{cc-inter} augmented with a temporal component using domain-specific time masking as described in Section \ref{sssec:tm-d}.
}
\label{fig:model-diagram}
\end{center}
\end{figure*}

An encoder-decoder model is used to evaluate each slot candidate, as shown in Figure~\ref{fig:model-diagram}. The current turn, past user turns, and past system turns are all encoded using an LSTM layer with attention. Each slot (key and value) and intent are also encoded by averaging the word embeddings contained in each. Finally, the \textit{slot distance} is encoded by counting the number of turns back that the slot appeared in (this would equal zero for slots from the current turn) and one-hot encoding that value. This is shown by the "Recenty One-Hot" input in the diagram. The final encoded slot candidate is passed to the decoder which produces a final carryover probability that determines whether or not the slot should be carried over to the current turn\footnote{The inputs shown in red are not part of the original formulation in~\citep{cc-inter}.}.

\subsection{Simple Time Mask (STM)}
\label{sssec:tm-s}

Inspired by~\cite{time-mask}, we introduce the concept of \textit{masked} embeddings so that irrelevant dimensions are suppressed in the embedding of the slots. 
We start by constructing a \textit{time embedding} based on the \textit{temporal distance}, $d_{\Delta t}$, of each candidate slot\footnote{
Defined as number of seconds in the past that the turn which contains the slot occured relative to the current utterance.}.
This is shown in Fig.~\ref{fig:model-diagram} as the bottom input, in the red box.
The time embedding is given by
\begin{eqnarray}
\label{eq:time-embedding}
\bm{d_t}=\phi(W_t d_{\Delta t} + \bm{b_t}),
\end{eqnarray}
where $\bm{d_t}$ is a nonlinear transformation implemented as a single layer feedforward neural network with weight matrix $W_t \in \mathbb{R}^{N_t \times 1}$ and $N_t$ is 
dimensionality of the time embedding vector. The time mask, $\bm{m_t}$, is computed by passing the time embedding, $\bm{d_t}$, through another feedforward neural network
\begin{eqnarray}
\label{eq:time-mask}
\bm{m_t}=\sigma(W_{d_t} \bm{d_t} + \bm{b_{d_t}}),
\end{eqnarray}
where $W_{d_t} \in \mathbb{R}^{N_s \times N_t}$ and $N_s$ is the dimensionality of the candidate slot embedding, $\bm{h_s}$.

Finally, we apply the time mask to the encoded slot embedding:
\begin{eqnarray}
\label{eq:masked-slot-embedding}
\bm{h_s'}=\bm{h_s} \odot \bm{m_{d_t}},
\end{eqnarray}
The updated candidate slot embedding, $\bm{h_s'}$, is now passed to the decoder in the exact same way as in the baseline slot carryover model as described in~\cite{cc-inter}.

\begin{figure}[htb]
\centering
\centerline{
    \includegraphics[scale=0.34]{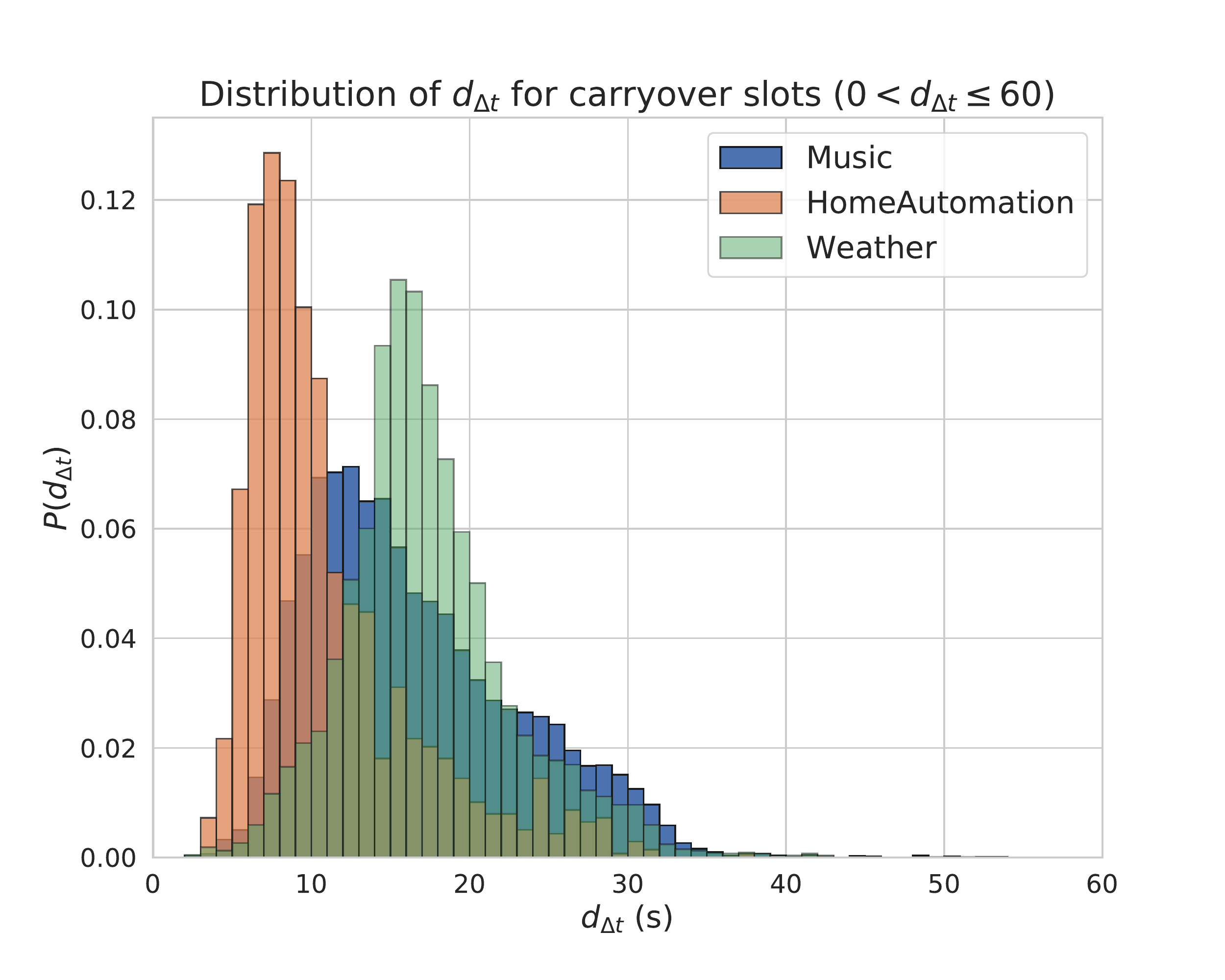}
}
\caption{Distribution of temporal distance, $d_{\Delta t}$, for all carryover slots from three of the largest domains in the IPDA dataset.}
\label{fig:domain-dist}
\end{figure}

Temporal dialogue behavior can vary by domain. Figure~\ref{fig:domain-dist} shows how much the distribution of $d_{\Delta t}$ can differ between three different domains in an internal IPDA dataset (described in more detail in ~\ref{ssec:dataset}). Therefore, we consider two extensions of the time masking approach that take into account the multi-domain nature of IPDAs.

\subsubsection{Intent Specific Time Mask (ITM)}
\label{sssec:tm-i}

We leverage the dialogue act or intent associated with the current turn in the time mask model.
In this formulation the time embedding is now given by
\begin{eqnarray}
\label{eq:time-embedding-i}
\bm{d_t}=\phi(W_t \bm{d_{\Delta t, a}} + \bm{b_t}),
\end{eqnarray}
where $\bm{d_{\Delta t, a}}=d_{\Delta t} \oplus \bm{h_a}$ is just the temporal distance concatenated with the existing intent embedding $h_a$ and now $W_{t} \in \mathbb{R}^{N_t \times (N_a + 1)}$, where $N_a$ is the number of dimensions used in the intent embedding.

\subsubsection{Domain Specific Time Mask (DTM)}
\label{sssec:tm-d}

We also try more coarse-grained, domain-level information as input to the time embedding.
Here we use a one-hot encoded representation of the domains, which gives us:
\begin{eqnarray}
\label{eq:time-embedding-d}
\bm{d_t}=\phi(W_t \bm{d_{\Delta t, D}} + \bm{b_t}),
\end{eqnarray}
where $\bm{d_{\Delta t, D}}=d_{\Delta t} \oplus \bm{1_D}$ is the concatenation of the temporal distance with the one-hot-encoded domain, $\bm{1_D}$, and $W_{t} \in \mathbb{R}^{N_t \times (N_D + 1)}$, where $N_D$ is the number of dimensions used in the one-hot-encoded Domain representation.
This architecture is shown in in Figure~\ref{fig:model-diagram}.

\subsubsection{Time-Decay Attention (TDA)}
\label{sssec:tda}

For comparison, we re-implemented the time-decay attention (TDA) model introduced in~\cite{su2018time}. However, the original work does not actually use time as a feature input but rather the \textit{ordinal distance} (equivalent to slot distance in our formulation) of each dialogue turn from the current utterance. To compare with our methods we use the actual temporal difference between dialogue turns in our implementation of the TDA model. The parameters are learned in the end-to-end training process.

\section{Experiments}
\label{sec:experiments}
\subsection{Dataset}
\label{ssec:dataset}
We present the results on 2 datasets. The {\bf IPDA dataset}, in  Table~\ref{tbl:data}, is an internal benchmark dataset collected from an IPDA for the en-US locale based on real usage. It consists of interactions over 7 domains - Music, Weather, LocalSearch, SmartHome, Video, MovieShowTimes, and Question Answering.
The data is transformed into individual candidate slots that are presented to the model, which determines whether or not they are relevant for the given turn.
For benchmarking against a public corpora, we also measure performance on DSTC2~\cite{henderson2014second} dataset.
We post-process the dataset similar to the internal dataset and only consider the top ASR and SLU hypothesis in addition to the system turn, dialogue acts and the associated slots.

\begin{figure*}[ht]
\centering
\subfigure[IPDA dataset.]{
    \includegraphics[scale=0.33]{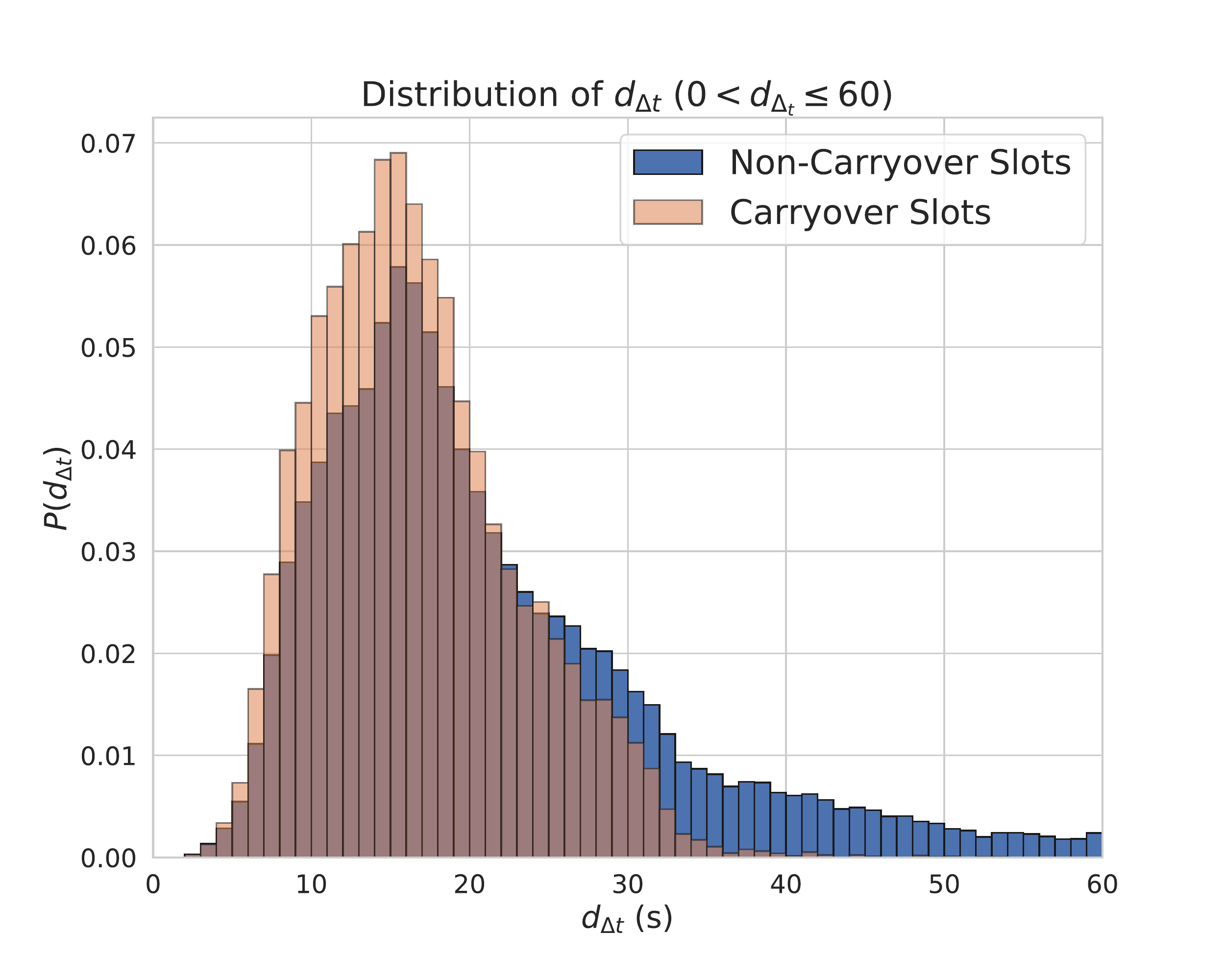}
    \label{fig:subfig1}
}
\subfigure[DSTC2 dataset.]{
    \includegraphics[scale=0.33]{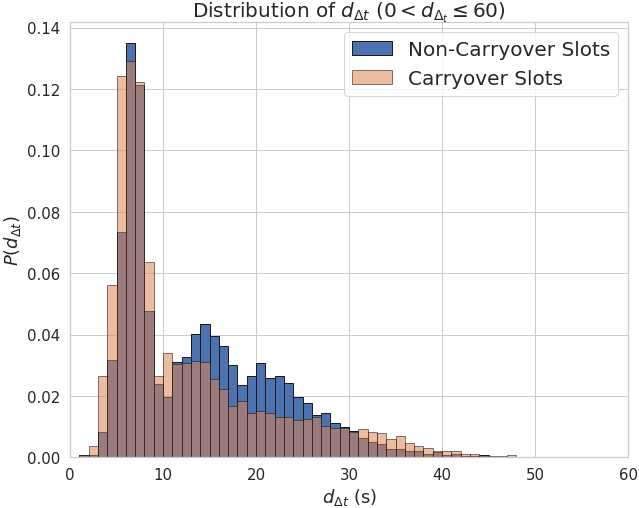}
    \label{fig:subfig2}
}
\vspace{-0.3cm}
\caption{
Distribution of temporal distance, $d_{\Delta t}$, for all candidate slots. The ``Carryover Slots", shown in orange, represent slot candidates found in the context that \textit{should} be carried over to the current turn.
}
\label{fig:pos-neg-dist}
\end{figure*}

\begin{table}[ht]
\small
\begin{center}
\begin{tabular}{|c||c|c|c|} \hline
& Train & Dev & Test \\\hline
Total & 264148 & 32437 & 33747 \\\hline
Positive Carryover & 92084 & 11389 & 11769 \\\hline
Avg. $d_{\Delta t}$ & 15.33s & 15.58s & 15.31s \\\hline
\end{tabular}
\end{center}
\caption{
IPDA dataset statistics. Here 'positive carryover' slots is the number of candidate slots that are relevant for the current turn.
}
\label{tbl:data}
\end{table}
Figure~\ref{fig:pos-neg-dist} shows the distribution of time between turns for both datasets.
If a slot candidate came from a context turn that was spoken 20 seconds before the current turn then $d_{\Delta t}=20$.
Based on human judged ground-truth, the slots that should be carried over to the current turn are shown in orange and the slots that should not are shown in blue.
One clear difference between the two distributions in the IPDA dataset is the long tail of the non-carryover distribution, indicating carryovers are more likely from a recent turn.
The domain specific distributions further indicate that leveraging dialogue time could be useful.

\subsection{Results on IPDA Dataset}
\label{ssec:results}
From Table~\ref{tbl:results-dt}, we can see that the TDA models offer a slight improvement over the baseline model.
Both models incorporate slot distance offset but the attention mechanism provides an additional boost.
The time mask models show additional gains demonstrating that leveraging dialogue time from each turn is important.
Moreover, the time information provides complementary information over the distance offset based measure, as shown by the improvements of the time masking models over the baseline model.
The DTM model performs the best overall in terms of F1, which suggests that adding domain information into the time mask provides additional disambiguation power.
Interestingly, we see that the ITM model does not improve much over the STM model, possibly because the intent embeddings do not necessarily distinguish between temporal behavior, and are already being leveraged by the slot carryover model. 

\begin{table*}[ht]
\small
\begin{center}
\begin{tabular}{|c||c|c|c|c||c|} \hline
Model & Overall & $d_{\Delta t}=(0, 15]$ & $d_{\Delta t}=(15, 30]$ & $d_{\Delta t}=(30, 60]$ & Overall DSTC2 F1 \\\hline
Baseline~\citep{cc-inter} & 87.8 & 89.3 & 86.8 & 74.3 & 95.0\\
STM & 88.4 & 89.7 & 87.5 & 77.5 & \textbf{96.1} \\
ITM & 88.6 & 89.8 & 87.8 & 76.9 & - \\
DTM & \textbf{89.2} & \textbf{90.5} & \textbf{88.3} & \textbf{80.0} & - \\\hline
TDA~\cite{su2018time}   & 88.4 & 90.0 & 87.5 & 72.8 & 94.6 \\\hline
\end{tabular}
\end{center}
\caption{
Overall F1 scores on the IPDA and DSTC2 dataset as well as F1 scores binned by $d_{\Delta t}$ for the IPDA dataset, which is measured in seconds. Note: the DSTC2 dataset only contains a single domain 
}
\label{tbl:results-dt}
\end{table*}

\subsubsection{Investigating longer temporal distance}
\label{sssec:large-t}

Here, we investigate the ability of the models to maintain higher accuracy over longer time windows in the dialogue context.
The overall F1 scores for each model are binned by $d_{\Delta t}$ for each candidate slot.
The results are shown in Table \ref{tbl:results-dt}.
The domain specific time mask model performs the best in each $d_{\Delta t}$ bin.
The effect of adding dialogue time information significantly improves performance in the largest $d_{\Delta t}$ range.
This is likely due to the model learning that older slots are less relevant to the current turn, which is impossible for the baseline model to do.
Additionally, we can see that the TDA model performs comparably to the STM model in the range $0 < d_{\Delta t} \leq 30$ but in the highest bin ($30 < d_{\Delta t} \leq 60$) we see that it falls well short of all of the time-masked models.

\subsection{Results on DSTC2 Dataset}

Since there is only one domain in DSTC2, we chose to only implement the STM model.
From the last column in Table~\ref{tbl:results-dt}, we can see that the STM model produces the best result.
The TDA model, contrary to previously reported results on DSTC4, does not perform as well.
Our hypothesis is that the temporal distribution across turns is not monotonically decaying, which is an assumption made in their approach.

\section{Related Work}
\label{sec:related}
Previous work on leveraging temporal information for dialogue state tracking has focused mostly on using distance offsets.
~\cite{chen2017dynamic} presented a time-aware attention network to leverage both contextual and ordinal distance information (i.e. the number of turns back from the current turn) and saw significant improvement.
Subsequently,~\cite{su2018time} improved upon this by designing a more flexible data-driven time attention mechanism that applied continuously decaying weights to past utterances before being fed into a contextual encoder.
The attention weight was determined based on the distance offset relative to the current turn.
However, distance offset is unable to capture complex dialogue scenarios, and our work improves upon this by modeling the actual wall-clock time difference between the current turn and the contextual turns.
This is particularly important in a multi-domain setting where a few second pause between consecutive user turns can be interpreted very differently depending on the dialogue scenario, and our experiments support our hypothesis.

Embedding masks have been explored in machine translation.
~\cite{context-mask} showed that contextualized word embeddings could be constructed from static word embeddings by applying a learned context mask.
This context mask allows the word to have different representations depending on the source sentence context around the word that is being translated, and the authors demonstrated improvements in machine translation tasks with this approach.
The approach of masking word representations was also explored in~\cite{ruseti2016using} for categorizing words into their wordnet classes.
We extend this masking concept to dialogue state tracking task, where we encode the temporal information in the dialogue as the masking operation over slots.

\section{Conclusion}
In this work we presented a novel approach for incorporating dialogue time information in multi-domain large-scale SLU systems.
We showed that our proposed time masking strategy provided gains over baseline systems that simply encode dialogue distance.
We presented several methods for incorporating additional information such as domain and intents into the time mask, and showed that this approach improved over competing approaches that indirectly incorporate time, particularly for multi-domain dialogues.
In the future, we want to investigate more contextualized representations of the domain and intent in order to capture more subtle variations in the dialogue for multi-domain scenarios.


\bibliographystyle{acl_natbib.bst}
\bibliography{cc_time_masked_attention}

\end{document}